\documentclass[manuscript, screen]{jair}
\hypersetup{hidelinks}

\setcopyright{cc}
\copyrightyear{2025}
\acmDOI{10.1613/jair.1.xxxxx}

\JAIRAE{Sven Koenig}
\JAIRTrack{} 
\acmVolume{4}
\acmArticle{6}
\acmMonth{8}
\acmYear{2025}

\RequirePackage[
  datamodel=acmdatamodel,
  style=acmauthoryear,
  backend=biber,
  giveninits=true,
  uniquename=init,
  hyperref=false
  ]{biblatex}

\usepackage{graphicx}
\usepackage{caption}
\usepackage{subcaption} 
\usepackage{amsmath}
\usepackage{algorithm}
\usepackage[noend]{algorithmic}
\usepackage{booktabs}
\usepackage[final,commandnameprefix=ifneeded]{changes}
\definechangesauthor[name={Revision Author}, color=purple]{rev1}
\definechangesauthor[name={Revision Author}, color=blue]{rev2}

\newcommand{\RightComment}[1]{%
    \hfill{\texttt{\footnotesize //#1}}%
}
\usepackage{bold-extra}
\usepackage{array}
\usepackage[acronym]{glossaries-extra}
\setabbreviationstyle[acronym]{long-short}
\setabbreviationstyle[short]{short-nolong}
 
\newacronym{MAT3C}{\textsc{mat3c}}{Multi-agent Pathfinding under Team-Connected Communication Constraint}
\newacronym{COMP}{\textsc{comp}}{\textit{centralized planning with composite state}}
\newacronym{PLF}{\textsc{plf}}{\textit{platooning leader-follower}}
\newacronym{PBSCOMM}{\textsc{pbs-comm}}{\textsc{pbs} under \gls{TCOMM}}
\newacronym{PIBTCOMM}{\textsc{pibt-comm}}{\textsc{pibt} under \gls{TCOMM}}
\newacronym{ODIDCOMM}{\textsc{odid-comm}}{\textsc{od-id} under \gls{TCOMM}}
\newacronym{MAAPGDL}{\textsc{apedl}}{Multi-Agent Pathfinding with Adaptive Path Expansion and Dynamic Leading}

\newacronym{APG}{\textsc{ape}}{\textit{adaptive path expansion}}
\newacronym{LOS}{\textsc{los}}{\textit{line-of-sight communication}}
\newacronym{ACOMM}{\textsc{acomm}}{\textit{agents communication constraint}}
\newacronym{TCOMM}{\textsc{tcomm}}{\textit{team communication constraint}}
\newacronym{PBS}{\textsc{pbs}}{\textit{priority-based search}}
\newacronym{PIBT}{\textsc{pibt}}{\textit{Priority inheritance with backtracking}}
\newacronym{MAPF}{\textsc{mapf}}{\textit{multi-agent pathfinding}}
\newacronym{TCT}{\textsc{tct}}{\textit{team communication tree}}
\newacronym{SAPFDL}{\textsc{sapf-dl}}{Single-Agent Pathfinding with Dynamic Leading}
\newacronym{LCR}{\textsc{lcr}}{\textit{limited communication range}}
\newacronym{DBS}{\textsc{dbs}}{\textit{deadlock-based search}}
\newacronym{CCBS}{\textsc{ccbs}}{\textit{continuous conflict-based search}}
\newacronym{CBS}{\textsc{cbs}}{\textit{conflict-based search}}
\newacronym{DL}{\textsc{dl}}{\textit{dynamic leading}}
\newacronym{OTIMAPP}{\textsc{otimapp}}{\textit{offline time-independent multi-agent path planning}}
\newacronym{MAPP}{\textsc{mapp}}{\textit{multi-agent path planning}}
\newacronym{BMAASTAR}{\textsc{bmaa}*}{\textit{bounded multi-agent \textsc{a}*}}
\newacronym{OD}{\textsc{od}}{\textit{operator decomposition}}
\newacronym{ID}{\textsc{id}}{\textit{independent detection}}

\newcommand{\malcr}{{\gls{MAT3C}}}
\newcommand{\planner}{{\gls{MAAPGDL}}}
\newcommand{\saplannerdl}{{\gls{SAPFDL}}}
\newcommand{\acomm}{{\gls{ACOMM}}}
\newcommand{\tcomm}{{\gls{TCOMM}}}
\newcommand{\apg}{{\textit{adaptive path expansion}}}
\newcommand{\tctree}{{\Symbol{T}$_{tc}$}}

\newcommand{\Acronym}[1]{\ensuremath{{{\texttt{#1}}}}}
\newcommand{\Symbol}[1]{\ensuremath{\mathcal{#1}}}
\newcommand{\Function}[1]{\ensuremath{{\scriptstyle{{\textsc{#1}}}}}}
\newcommand{\Var}[1]{\ensuremath{{{\mathrm{#1}}}}}
\newcommand{\False}{\ensuremath{\Acronym{false}}}
\newcommand{\True}{\ensuremath{\Acronym{true}}}

\newcommand{\Subdiv}{\ensuremath{\Delta}}
\newcommand{\World}{\Symbol{W}}

\newcommand{\Obstacle}{\Symbol{O}}

\newcommand{\GoalPos}{\ensuremath{g}}
\newcommand{\Traj}{\ensuremath{\zeta}}

\newcommand{\Tree}{\Symbol{T}}

\begin{document}

\title[MAPF under T3C via Adaptive Path Expansion and Dynamic Leading]{Multi-Agent Pathfinding Under Team-Connected Communication Constraint via Adaptive Path Expansion and Dynamic Leading}

\author{Hoang-Dung Bui}
\authornote{Corresponding Author.}
\orcid{0000-0003-1901-7120}
\email{hbui20@gmu.edu}
\affiliation{%
  \institution{George Mason University}
  \city{Fairfax}
  \state{Virginia}
  \country{USA}
}

\author{Erion Plaku}
\orcid{0000-0002-6622-386X}
\email{eplaku@nsf.gov}
\affiliation{%
  \institution{U.S. National Science Foundation}
  \city{Alexandria}
  \state{Virginia}
  \country{USA}  
}

\author{Gregory J. Stein}
\orcid{0000-0003-1981-4154}
\email{gjstein@gmu.edu}
\affiliation{%
  \institution{George Mason University}
  \city{Fairfax}
  \state{Virginia}
  \country{USA}
}

\renewcommand{\shortauthors}{Bui, Plaku, \& Stein}

\begin{abstract}
This paper proposes a novel planning framework to handle a multi-agent pathfinding problem under a team-connected communication constraint, where all agents must have a connected communication channel to the rest of the team during their entire movements.
Standard multi-agent pathfinding approaches (e.g., priority-based search) have potential in this domain but routinely fail when neighboring configurations at start and goal differ. Their single-expansion approach---computing each agent's path from the start to the goal in just a single expansion---cannot reliably handle planning under communication constraints for agents as their neighbors change during navigating.
Similarly, leader-follower approaches (e.g., platooning) are effective at maintaining team communication, but fixing the leader at the outset of planning can cause planning to become stuck in dense-clutter environments, limiting their practical utility.
To overcome this limitation, we propose a novel two-level multi-agent pathfinding framework that integrates two techniques: \textit{adaptive path expansion} to expand agent paths to their goals in multiple stages;
and \textit{dynamic leading} technique that enables the reselection of the leading agent during each agent path expansion whenever progress cannot be made.
Simulation experiments show the efficiency of our planning approach, which can handle up to 25 agents across five environment types under a limited communication range constraint and up to 11--12 agents on three environments types under line-of-sight communication constraint, exceeding 90 \% success-rate where baselines routinely fail. 

\end{abstract}


\received{22 July 2025}
\received[revised]{10 October 2025}
\received[accepted]{25 October 2025}

\maketitle

\section{Introduction} 
\label{sec:Intro}

We want a team of agents to navigate through an obstacle-rich environment to reach goals while maintaining constant team communication: a spanning tree created from range-limited or line-of-sight communication between pairs of agents.
This problem is relevant to scenarios such as supply delivery during disasters or monitoring hostile environments \added{, where maintaining constant communication can be a means of ensuring resiliency, allowing the team to immediately plan anew if the mission or environment were to change. There is thus a need for algorithmic solutions to this problem, which can quickly and effectively plan trajectories for an entire team to reach its target destinations while constrained to maintain team communication.}
\deleted{To mitigate this, agents must ensure that the team is in constant communication throughout their movement.} Maintaining the spanning tree while the agents head in different directions with varied lengths of actions makes pathfinding challenging in continuous time and space, even for holonomic agents. 
The challenge is compounded by agents having varying neighboring configurations between start and goal positions which limits the efficacy of selecting a single \emph{leader agent} that other agents can follow. Furthermore, agents must navigate through narrow passages and non-convex spaces without collisions to reach the goal configurations. Collectively, we refer this problem domain as \gls{MAT3C}.

The problem can theoretically be solved using composite state  approaches~\cite{wagner2015subdimensional,standley2010finding},
in which planning selects between team actions, each simultaneously specifying an action for all agents at once. 
However, this approach quickly suffers from the curse of dimensionality, as the difficulty of planning increases exponentially with the number of agents, making planning intractable for even relatively small problems.

\begin{figure}[t]
    \centering
    \includegraphics[width=0.8\columnwidth]{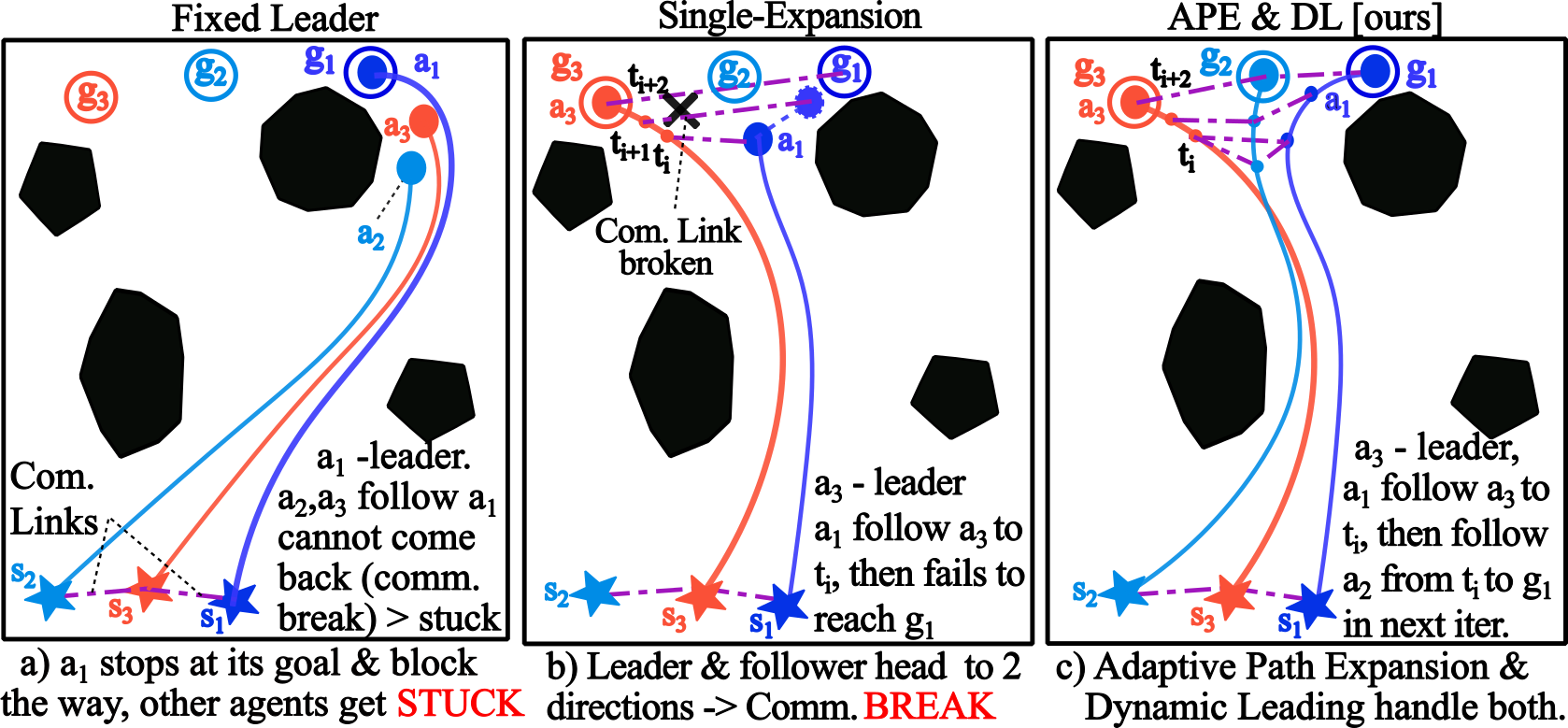}
    \caption{\textbf{Fixed leader and single-expansion of agent paths fail in scenarios a) \& b). Our proposed techniques: \textit{adaptive path expansion} and \textit{dynamic leading} handle both (c).} If $a_1$'s leading causes the team to get stuck (a), \textit{dynamic leading} allows another agent take over as the new leader. When the leader and followers move to different directions (b), \textit{adaptive path expansion} allows the stuck-agent to resume planning in next expansion.}
    \Description{This figure presents two scenarios (a,b) on which classical \textsc{mapf} fail to solve under team-connected communication constraint. Our proposed planner,\textsc{apedl}, can solve both.}
    \label{fig:Intro}
\end{figure}

The decoupled approach~\cite{solovey2016finding,sigurdson2018multi,okumura2023offline,okumura2022priority,andreychuk2022multi,andreychuk2021improving,ma2019searching} mitigates this challenge by planning each agent in sequence and resolving conflicts among paths by enforcing constraints and replanning.
However, incorporating the team communication constraint would require considerable algorithmic modifications to these planners, especially when an agents local neighbors must change between the start and goal configurations, as in Figure~\ref{fig:Intro}b.

Another approach is \textit{platooning leader-follower}~\cite{huang2019platoon,shojaei2019tracking,qian2016mpcautonomousvehicles,gao2019multilane} that handles the communication constraint by selecting a leader that has no communication constraint and others act as followers, who plan so as to maintain communication with the agent that planned before them.
However, establishing a fixed planning order for the leader agent can result in issues during planning. These issues include the follower agents finding no action to follow the leader within the communication range or becoming stuck once the leader reaches its goal (Figure~\ref{fig:Intro}a), or the team needing to spread out for the followers to reach their goals while maintaining the communication constraint (Figure~\ref{fig:Intro}b).

To address these challenges, this work proposes a novel planning algorithm for solving  \malcr, ensuring a connected communication channel in the team at any time from start to goal positions. 
Our planning algorithm integrates two techniques: \apg{}, which extends agents' paths across multiple expansions as the agents' progress stalls or they encounter conflicts; and \textit{dynamic leading}, which allows a new agent to assume the leader role if expanding via the current leader cannot make progress, enabling further expansion of each single-agent trajectory.

From these techniques, we have developed a planner, called \gls{MAAPGDL}, for planning in multi-agent systems under team connected-communication constraint. We show our planner is effective under two types of communication constraint: \gls{LCR} and \gls{LOS}.  
Simulated experiments show that our planner results in fast and effective planning, handling up to 25 agents under \gls{LCR} constraint across five environment types, and up to 12 agents with Random Forest and Rings environments, and up to 3--10 agents for Office, Waves, and Maze environments under \gls{LOS} constraint with more than 90\% success-rate where baselines routinely fail. 

The paper is structured as follows: Section~\ref{sec:Intro} introduces the \gls{MAT3C} problem and its challenges. Section~\ref{sec:related-work} discusses existing approaches, their potential to address the problem, and the obstacles they face. Section~\ref{sec:problem-formulation} presents
the problem formulation.
Section~\ref{sec:over_contr} provides an overview of our contributions, while Section~\ref{sec:plan_alg} details the planning algorithm. Section~\ref{sec:expres} discusses the experiments and results, Section~\ref{sec:completeAna} analyzes the planner's completeness, and Section~\ref{sec:conclusion} evaluates our contribution and outlines future work.

\section{Related Work} \label{sec:related-work}

There exist multiple approaches that seek to solve multi-agent pathfinding problems, in the absence of a communication constraint, that could in theory be adapted to solve the \malcr{}  problem.
To effectively scale to more agents, approaches in this space are typically designed around reducing the exponential growth of the planning tree with planning time horizon.
One class of planners limits state space expansion by introducing composite states only if necessary. Standley~(\citeyear{standley2010finding}) proposed two techniques: \gls{OD} reduces the branching factor by allowing only one agent to select an action at each timestep; and \gls{ID}, a collision-checking and replanning mechanism that determines whether composite states are required for planning.
Wagner and Choset~(\citeyear{wagner2015subdimensional}) introduced the \textsc{m}$^*$ algorithm, which used \textit{individual policies} to guide agents for next actions 
if no collision occurs, 
yet builds composite states for agents by \emph{collision set} when their actions are in collision.
Though \textsc{m}$^*$ and \textsc{od-id} algorithm work well if the agents are far apart, when agents are close to each other as in the \malcr{} problem, collisions occur frequently, causing the planners eventually to use joint-state and struggle to handle the exponential growth of joint-state space; thus limiting the efficacy of these approaches for solving our limited-communication problem, which requires the agents to be close to one another.  

Another class of planners uses the decoupled approach that plans each agent individually, deploying additional techniques to resolve conflicts among agents and replanning as needed.
Andreychuck et al.~(\citeyear{andreychuk2022multi,andreychuk2021improving}) developed \gls{CCBS}---a variant of \gls{CBS} in continuous time---that defines constraints with time period for conflicting agents.
Ma et al.~(\citeyear{ma2019searching}) proposed \gls{PBS} that searches for a priority of planning order among agents, which results in conflict-free among the single agent paths.
\citet{okumura2023offline} proposed the \gls{OTIMAPP} planner, which employs \gls{DBS} to resolve conflict among paths determined through prioritized planning.
Although \gls{CCBS}, \gls{PBS} and \gls{OTIMAPP} are robust to compute paths for standard \gls{MAPF} problem, their approaches of single-expansion for agent paths---computing each agent's path from the start to the goal in just one expansion---fail to solve \gls{MAT3C} problem in Figure~\ref{fig:Intro}b scenario.

\gls{PIBT}~\cite{okumura2022priority} and \Gls{BMAASTAR}~\cite{sigurdson2018multi} expand a single action for each agent in each planning iteration. These planners can handle well deadlocks among the agents. However, there are two issues for these planners to solve the \malcr{} problem. The first one is their myopic action 
selection strategy, which can result in situations where followers are unable to find actions that keep them within the leader's communication range. The second one is their fixed planning order causing followers to fail to plan when their goal positions are out of communication with the leader's goal position (Figure~\ref{fig:Intro}b).
Wang and Botea~(\citeyear{wang2011mapp}) introduced the \textsc{mapp} planner, which requires that sufficient space exists for agents to trade places with one another to guarantee completeness, and also does not consider communication restrictions.
In addition to the limitation of a fixed planning order (as mentioned above), \textsc{mapp} would require additional algorithmic changes to modify the computed paths while adhering to deadlock and communication constraints.

Choudhury et al.~(\citeyear{choudhury2022scalable}) proposed an online \textsc{fv-mcts-mp} planner, which utilized coordinate graphs and the max-plus algorithm to address the challenge of action space growth as the number of agents increases. However, in the \malcr{} problem, neighbors are dynamic, and the agent team only needs to maintain a spanning tree rather than a fully connected graph. These features complicate the setup of local payoff components, limiting the planner's efficiency in solving the \malcr{} problem and needlessly restrict the space of possible solutions.
Solovey, Salzman, and Halperin~(\citeyear{solovey2016finding}) introduced the d\textsc{rrt} planner, which uses an implicit composite roadmap---a product of single-agent roadmaps. Their algorithm grows a multi-agent tree in this composite roadmap using \textsc{rrt}. However, using such a roadmap requires abstracting low-level 
details of each robot's motion over time, making it difficult to apply their approach for planning with communication constraints.
In general, it is very challenging and not straightforward to adapt these approaches to solve the \malcr{} problem
in which each member of the team must be in constant communication with one another in continuous time. 

Several algorithms~\cite{bhattacharya2010multi,diaz2017general,pal2012multi} were developed for limited versions of the team communication constraint. Bhattacharya et al.~(\citeyear{bhattacharya2010multi}) employed a soft constraint with incremental penalties to enforce communication constraint on pairs of agents at specific predefined points en route to goals. However, the \gls{MAT3C} problem requires the agents to maintain continuous team communication. It is also challenging to apply soft constraints on the entire agent team because the constraints of later agent pairs can break the communication constraints of earlier pairs. 
Pat et al.~(\citeyear{pal2012multi}) introduced a planning framework for exploration, but their team communication constraint applies only at agents' frontiers and goals, not during movement.
To the best of our knowledge, no \gls{MAPF} literature addresses the \gls{MAT3C} problem.

Formation control~\cite{sehn2024hard,kowdiki2019autonomous,garrido2013general,qian2016mpcautonomousvehicles,aljassani2023enhanced} is a straightforward method to address the \malcr{} problem if relaxing the formation requirement. In this approach, leader agents are planned first, followed by the paths for follower agents, which maintain predefined formations while following the leaders that ensure constant communication among the agents. 
However, this approach encounters challenges when the relative positions of the agents at start and goals are randomly arranged.

Specific to the \gls{MAT3C} problem, \emph{platooning} has emerged as a state-of-the-art solution~\cite{shojaei2019tracking} in which the motion of a leader-agent is planned first and follower-agents planned one at a time in sequence to follow the agent that preceded them, and so is designed to maintain communication constraints between pairs of agents.
Most existing work in platooning performs full motion planning only for the leader and uses a low-level controller to regulate followers to maintain a distance from the preceding agents~\cite{shojaei2019tracking,huang2019platoon,agachi2024rrtlf}.
Zhao et al.~(\citeyear{zhao2017design}), and Gao et al.~(\citeyear{gao2019multilane}) 
introduced a model predictive controller that generates trajectories for a virtual center, which followers need to track.
However, recent work in the space of \gls{CBS}~\cite{ma2019searching} 
has proven that planning with a fixed vehicle order in general, including platooning, is incomplete and frequently becomes stuck when the leader moves in a direction different to the follower's goals, as we show in Figure~\ref{fig:Intro}a. 

To overcome the limitations of using a fixed leader in a planning iteration and handling neighbor changing as navigating from start and goal, a planner should integrate two key capabilities: \emph{dynamic leading}, in which the leader can be changed in a single-agent path expanding when the team fails to make progress towards the goals; and \emph{adaptive path expansion}, that allows an agent to resume its planning in next path expansions if expansion stalls.

\section{Problem Definition} \label{sec:problem-formulation}

We formally define the \gls{MAT3C} problem as follows.
There are $n$ agents $\{a_1, \ldots, a_n\}$ and a known map of the world $\World$ with obstacles $\Obstacle = \{\Obstacle_1, \ldots, \Obstacle_o\}$. 
The agents start at the initial positions $s^\Var{init} = \{s^\Var{init}_1, \ldots,s^\Var{init}_n\}$ and 
head to the goals $\GoalPos = \{\GoalPos_1, \ldots,\GoalPos_n\}$, where $s_i, g_i \in \World$ and $s_i, g_i \notin \Obstacle$. 
The configurations of initial positions and goals are chosen randomly but such that they still satisfy a team communication constraint, defined below.

The world $\World$ is divided into sub-divisions $\Subdiv$, which are obstacle-free regions. 
A graph $\mathcal{G}$ is created where the vertices are the centroids of $\Subdiv$ and the edges represent connections of neighboring sub-divisions. 
An agent moves to a neighboring sub-division with constant velocity $\Var{v}_c$ by taking the action: \textit{Move}$(v,u)$, where $v,u$ are neighbor vertices in $\mathcal{G}$. 
As reaching its goal, the agent takes the action of \textit{Stop} = \textit{Move}$(v,v)$ \added{and comes to rest at its goal location. Due to the communication constraint, subsequent agents still need to check collision with agents at their goal locations.} 
Agents moving over $\mathcal{G}$ are guaranteed to be \textit{collision-free} with respect to the static obstacles.

We define two levels of communication constraint for the agents. 
The first level is between two agents, referred to as \gls{ACOMM}. \added{In this work, we consider two different means of establishing communication between two agents: (1) the \glsentryfull{LCR} constraint: if the Euclidean distance between the two agents’ positions is less than or equal to the communication range $r_c$, and (2) the \glsentryfull{LOS} constraint: the two agents have an unobstructed line of sight to each other.}
The second level applies to the entire agent team, referred to as \gls {TCOMM}, which requires the team to form a spanning tree where the edges represent the connections between pairs of agents satisfying the \gls{ACOMM} constraint.
An action \textit{Move($v,u$)} satisfies the \gls{ACOMM} constraint 
if during the movement \textit{Move($v,u$)} the agent has at least one 
neighboring agent within a distance of $r_c$.
Due to varied lengths of actions and \added{\gls{TCOMM} requirement}, time and agents' positions are continuous in \gls{MAT3C}\added{, even though the actions are discrete}.

A \textbf{collision} occurs between two agents, $a_i$ and $a_j$, when their distance is less than a threshold $d_c$ at timestep $t$. 
A \textbf{path} is a sequence of waypoints to transition an agent from a position $p_s$ to position $p_g$ with constant velocity $\emph{v}_c$. We say a path is \textbf{reach-goal} if it can lead the agent to the goal, is collision-free, and the movement between two sequential waypoints satisfies \gls{ACOMM} constraint;
When an agent stops at its goal, we still consider collision and \gls{ACOMM} constraints. 
A path is \textbf{valid} if it is \emph{reach-goal} and  after reaching the goal at time $t_g$ forever stays at the goal, satisfies \gls{ACOMM}, and is collision-free.

The objective is to compute \textit{valid} paths $\{\Traj_1, \ldots, \Traj_n\}$, one for each agent, so that $\Traj_i$ starts at $s^\Var{init}_i$, reaches $\GoalPos_i$, and the agents satisfy \gls{TCOMM} constraint from the start time to the time the last agent reaches its goal. 
We develop a \gls{MAPF} planner that seeks to reduce the overall planning time and the travel distances.

\section{Overview of Contributions} 
\label{sec:over_contr}
To address the \gls{MAT3C} problem, we need to solve several challenges, including maintaining a team communication channel at any time, changing of neighbors as navigating, and planning getting stuck due to the leader agent stalling. We propose two methods, \gls{APG} and \gls{DL}, to resolve these challenges. This section is organized as follows: Section~\ref{sec:meap} introduces \gls{APG}---that allows multistage growing of agent paths; Section~\ref{sec:dl} presents \gls{DL}, which enables seamless leader transitions within each single agent path expansion, ensuring planning progress with minimal reliance on planning-order and leader selection. Section~\ref{sec:method_tct} discusses a \gls{TCT} that manages planning progress and ensures the leader agent maintains team communication in case some \added{of its neighbor} agents have already reached their goals.

\subsection{Adaptive Path Expansion} 
\label{sec:meap}
In \added{our} \gls{MAT3C} \deleted{problem}\added{experiments}, the start and goal positions are randomly chosen while adhering to the \gls{TCOMM} constraint. Agents have different neighbors at the start and the goal, \deleted{causing to change their neighbors}\added{forcing them to rearrange and change their nearest neighbors} as they navigate from the start to the goal positions. 
When an agent follows the leader---who by definition is not affected by the \acomm{} constraint---to a branching point where both head to different goals, the follower agent will stall because no action satisfies the \acomm{} constraint (Figure~\ref{fig:Intro}b). 
If deploying single-expansion of agent paths (computing each agent's path from the start to the goal in a single expansion) as in \gls{PBS}, \gls{CCBS}, \textsc{od-id}, \gls{OTIMAPP}, or \textit{platooning} approaches, the planning fails because no feasible path can be found during the current expansion (Figure~\ref{fig:Intro}b). 
If each agent selects a single action based on its own interests in each iteration, as in \gls{PIBT} and \textsc{bmaa}$^*$, it would lead to situations where a follower cannot find a feasible action to maintain communication constraint with its leader (e.g. due to obstacles), resulting in failure of planning.

To address this issue, we introduce \glsentryfull{APG}: a method that enables agents to iteratively refine and expand their paths multiple times if they fail to reach their goals in prior expanding attempts. 
\added{The term \textit{adaptive} refers to two features: (1) that the number of path expansions can vary for each agent, with additional attempts to expand the plan potentially resolving conflicts borne of communication constraint violations, and (2) that the planned paths can be modified if \textit{collision-at-goal} situations occur in which an agent coming to rest at its goal location impedes progress of other agents along their own planned paths \cite{bui2024multi}.}
When expanding, \added{different from \gls{PBS} and \gls{OTIMAPP}, the planned paths in our planning framework must  satisfy the \gls{ACOMM} constraint with prior planned paths}. \added{As \textit{collision-at-goal} situations happen,}
rather than replanning the entire paths, we refine the prior planned paths by trimming them to the point of collision and then re-expanding from that point. This approach avoids breaking the communication constraints of follower agents, which would otherwise necessitate replanning for all subsequent agents. The \gls{APG} alone does not fully address the \gls{MAT3C} problem if the leader stalls or achieves goals before the follower agents. To resolve this, we propose the second technique \textit{dynamic leading}.

\subsection{Dynamic Leading}
\label{sec:dl}
State-of-the-art \gls{MAPF} planners, such as \gls{PBS}, \gls{OTIMAPP}, and \gls{PIBT}, allow the planning order to be modified before each planning iteration but keep it fixed during the iteration. This fixed planning order with the first agent as the leader can lead to failures in solving the \gls{MAT3C} problem as followers must continuously communicate with the leader, causing them to stall if the leader halts (Figure~\ref{fig:Intro}a). 

To address the issue, we propose \glsentryfull{DL}, a technique that dynamically reassigns the leader role to the agent that has progressed the farthest for guiding the team within the \textit{same planning order} during single-agent path expansion.
Choosing a new leader based solely on position or proximity to goals can cause the team to become stuck, especially when the current leader reaches its target.
The \gls{DL} method allows an agent to take the leader role if it has made the most temporal progress in planning. Leader switching occurs when the planning agent takes an action at time $t$ while other agents have not yet reached $t$, allowing the planning agent to assume the leader role and planning the action without adhering to the \gls{ACOMM} constraint.
Experiments demonstrate that the \gls{DL} technique performs efficiently, even with randomized selection of planning orders and leaders.

\begin{figure}[t]
  \centering
  \includegraphics[width=1.0\textwidth]{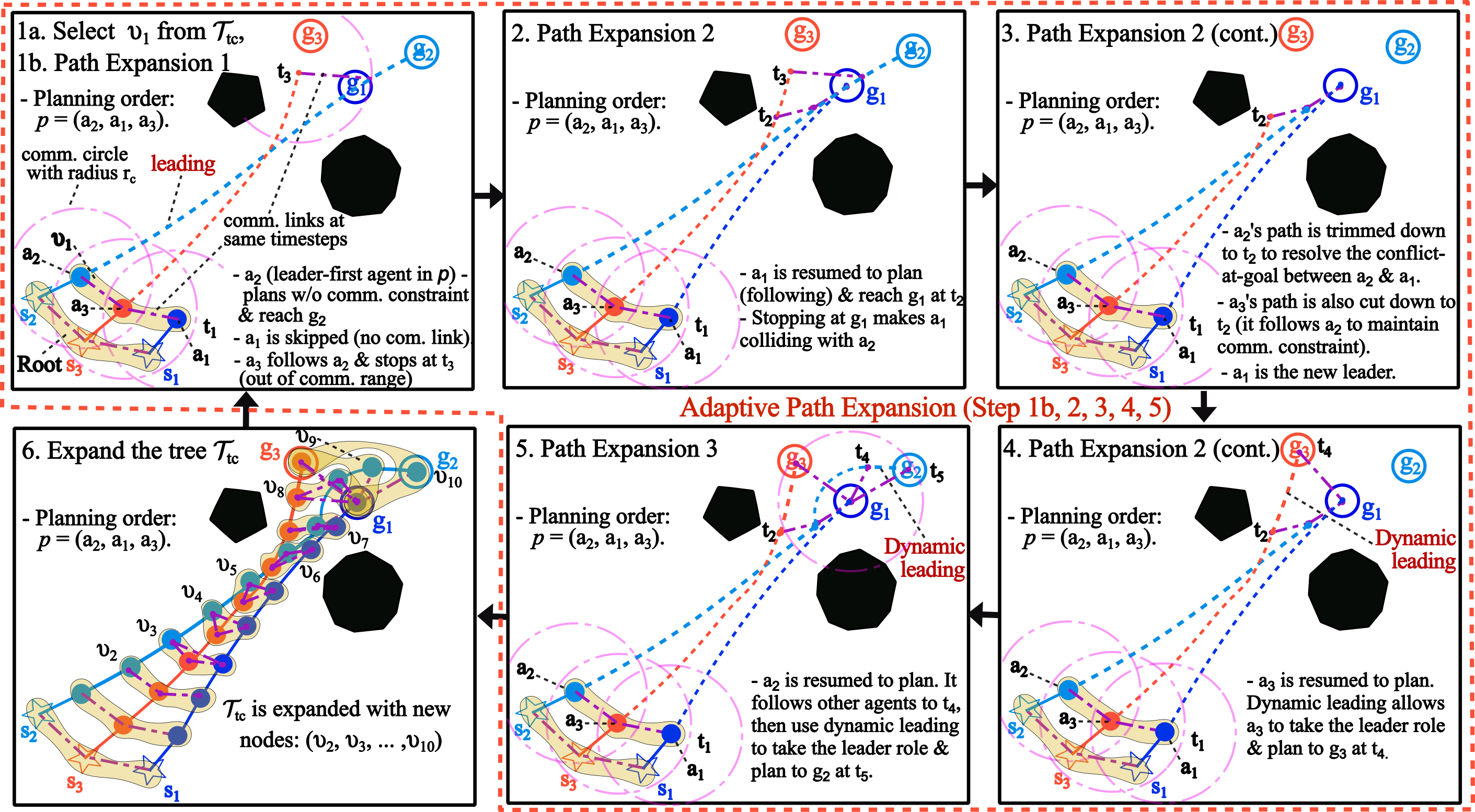}
  \caption{\textbf{An illustration of one-step \gls{TCT}'s expansion with 3 agents}. Agents $a_1$, $a_2$, and $a_3$, start at $s_1$, $s_2$, and $s_3$, and target their goals $g_1$, $g_2$, and $g_3$ while avoiding the obstacles (black) and other agents. From node $v_1$ in \tctree{} (step 1a), the planner expands trajectories for each agent initially with a planning order. Our first technique, \textit{adaptive path expansion}---from step 1b to 5, grows and refines agent paths. The second technique, \textit{dynamic leading}, at step 4 and 5, allows $a_3$ and $a_2$ to reach their goals.
  The tree \tctree{} is expanded at step 6.}
  \Description{This figure demonstrates our planning framework that consists of six steps to expand the team communication tree.}
  \label{fig:fw}
\end{figure}

\subsection{Team Communication Tree}
\label{sec:method_tct}

\textit{Dynamic leading} allows to switch the leader that can move without \gls{ACOMM} constraint. However, the new leader may expand to a new direction that disrupts communication with agents already at their goals. To address this challenge, we introduce a \textit{team communication tree} (\gls{TCT}) to manage planning progress \added{and avoid recomputing the entire agent paths from the starts}. If the leader's actions break the team's communication channel, the paths under its leadership are excluded from the tree. \added{Instead of computing the paths from the starts, the \gls{TCT} tree allows the planner to recompute from the closest node on the} \gls{TCT} for replanning.
The \gls{TCT} comprises agent paths, with nodes containing the agent locations, a timestep, an initial $f$-value assigned as follows:
\begin{equation}
\label{Eq:fvalue}
    f = \alpha g + (1- \alpha) h~,
\end{equation}
where $g$ and $h$ are the sum of $g_i$ and $h_i$ values from all agents. $\alpha$ is a weight factor that balances the trade-off between prioritizing shorter paths and faster goal convergence ($\alpha = 0.1$ in our experiments).

A node is added into the tree, if the agent's positions \added{at the node} satisfy the \tcomm{} constraint. An agent's position at a given time is interpolated from its path based on time. Due to continuous time \added{under the \tcomm{} constraint} and varied \added{length of actions, the interpolated agent's position can fall anywhere along its paths, and thus anywhere within a sub-division. Consequently, expanded paths can result in agent positions being in continuous-space locations throughout the workspace} workspace.
If a node is selected to expand (Alg.~\ref{alg:MADIST}.4), its $f$-value is penalized to encourage the planner to choose other nodes in the next iteration. 

From the proposed techniques, we develop the planner \gls{MAAPGDL} which combines all three of these techniques to to address the \gls{MAT3C} problem.

\section{Planning Framework}
\label{sec:plan_alg}
Our planning framework, which we call \textit{Multi-agent Pathfinding with Adaptive Path Expansion and Dynamic Leading} (\gls{MAAPGDL}), consists of two levels: High Level, which manages multi-stage growing of agents paths, and Low Level: \gls{SAPFDL}. 
\added{The major components of the framework are described as follows.}
Section~\ref{sec:MADIST} introduces the High Level (Alg.~\ref{alg:MADIST}, Figure~\ref{fig:fw})
\added{that consists of selecting a node from the tree (\gls{TCT}), expanding agent's paths and modifying them if necessary via \gls{APG}, then growing the tree until reaching the goals or running out of runtime}. Section~\ref{sec:SAPF} presents the Low Level: a single-agent pathfinding with dynamic leading (Alg.~\ref{alg:sapf}) that continuously expands the shortest collision-free and \acomm-satisfied paths to goals for the agents one by one.
Section~\ref{sec:TCT_alg} describes how the tree \gls{TCT} is expanded, and 
Section~\ref{sec:alg_analysis} analyzes the time complexity of the planner.

\begin{algorithm}[t]
  \caption{High Level: Growing a Team Communication Tree}
  \label{alg:MADIST}
  \begin{algorithmic}
  \STATE \text{INPUT}: $n$: \text{number of agents}, $g=\{g_1,...,g_n\}$: goals, $\World$: \Var{world}, $s^\Var{init}=\{s_1^\Var{init},...,s_n^\Var{init}\}$: \text{starts}, $t_\Var{ds}$: \text{max runtime}
  \STATE \text{OUTPUT: paths} $\Traj=\{\Traj_1,...,\Traj_\Var{n}\}$
  \end{algorithmic}
  \begin{algorithmic}[1]
  \STATE $\Tree_\Var{tc}\!\gets \Function{InitTCTree}(s^\Var{init}, t=0)$;
  $\Subdiv \! \gets\!\Function{SubDivision}(\World)$; $\mathcal{G} \gets \Function{CreateGraph}(\Subdiv, g)$\;
  \STATE $\mathcal{H} \gets \{\Function{CalHeursitic}(\mathcal{G}, g_1),\ldots, \Function{CalHeursitic}(\mathcal{G}, g_n)$\}\;
  \WHILE{$ \Function{time}() < t_\Var{ds}$}
      \STATE $v\! \gets\! \Function{SelectNode}(\Tree_\Var{tc})$; $p\! \gets\! \Function{RandomOrder}()$; $\Traj^s\! \gets\! \Function{InitPaths}(v)$ \RightComment{planning iter.(line:4-14)} 
      \FOR{$m \text{ iterations}$ \RightComment{adaptive path expansion (line:5-12)}}  
          \STATE {$\Var{allRG} \gets \True$}\; 
          \FOR{$1 \leq i \leq n$ \RightComment{single-expansions for agent paths (line:7-11)}} 
              \STATE \textbf{if } $\Function{FoundGoal}(\Traj^s_{p_i})$ \textbf{then continue}\; 
              \STATE $\Traj^t\!\gets\!\saplannerdl(p_i, g_{p_i}, \Traj^s, \mathcal{G}, \mathcal{H}_{p_i})$; $\Traj^s_{p_i}.\Function{Insert}(\Traj^t)$ \RightComment{a single-path expansion} 
              \STATE \textbf{if} $\Function{FoundGoal}(\Traj^s_{p_i})$ \textbf{then} $\Function{ModifyIfOverlap}(p_i, \Traj^s, \Var{allRG})$
              \STATE \textbf{else} $\Var{allRG}=\False$\; 
          \ENDFOR
          \STATE \textbf{if} $\Var{allRG}$ \textbf{then break}
      \ENDFOR
      \STATE $\Function{ExpandTCTree}(\Tree_\Var{tc}, \Traj^s, v)$\; 
      \STATE \textbf{if} $\Var{allRG}$ \textbf{then break}
  \ENDWHILE
  
  \STATE $\Var{vid} = \Function{ClosestNodeToGoal}(\Tree_\Var{tc})$\;
  \STATE \textbf{return} $\Function{GetPaths}(\Tree_\Var{tc}(\Var{vid}))$\;
  \end{algorithmic}
  \end{algorithm}

\subsection{High Level: Growing the Team Communication Tree with Adaptive Path Expansion}
\label{sec:MADIST}
The high level grows a \gls{TCT} with nodes composed of agent paths.
The nodes are collision-free and each have a configuration \added{of agent positions} that includes a spanning tree of communication among the agents. A node consists of a timestep, positions of agents, $f$-value \added{(defined by Eq.~\ref{Eq:fvalue})}.
The $f$-value is increased by a penalty factor when a node is reselected \added{(in \Function{SelectNode()} function at line Alg.~\ref{alg:MADIST}.4)} to encourage the planner to explore other nodes. 
The high-level module \added{develops the tree \gls{TCT} $\Tree_{tc}$ by the following steps: (1) selecting a node to expand, (2) randomizing the order in which to plan trajectories for the agents, (3) running the \gls{APG} loop, then (4) growing the tree from the returned paths}.

The module gets the number of agents $n$, the goals $g$, initial start $s^\text{init}$, and the world $\World$ as inputs and returns the \textit{valid} paths or, if valid paths cannot be found, those closest to the goals. 
The algorithm (Alg.~\ref{alg:MADIST}) starts by initializing a \gls{TCT} 
$\Tree_{tc}$ with the root consisting of all initial positions $s^\text{init}$ 
(Alg.~\ref{alg:MADIST}:1). 
The world $\World$ is divided into sub-divisions $\Subdiv$ of a predefined area, with obstacle-overlapping sub-divisions further subdivided 
along their largest dimension until they are clear or reach a size threshold.
A graph $\mathcal{G}$ is created from the centroids of the obstacle-free sub-divisions, while the edges represent the connections between neighboring sub-divisions (sharing boundaries or corners).
We compute \added{the shortest paths from all sub-divisions to each agent's goal $g_i$ used as a planning heuristic by the function $\Function{CalHeuristic()}$ and store them in the array $\mathcal{H}$ } (Alg.~\ref{alg:MADIST}:2)\added{, a one-time cost that greatly accelerates plan expansion.}.
The main loop (Alg.~\ref{alg:MADIST}:3--14) runs multiple planning iterations starting by selecting a node $v$ with smallest $f$-value 
(defined by Eq.~\ref{Eq:fvalue}) from $\Tree_{tc}$. \added{This $f$-value is then increased by factor $c$ ($c=0.05$ in our experiments) to encourage exploring other nodes}.
A planning order $p$ is created by a random function $\Function{RandomOrder()}$. 
A list of $n$ paths $\Traj^s$ is initialized from $v$ (Alg.~\ref{alg:MADIST}:4). 

The \glsentryfull{APG} technique (Alg.~\ref{alg:MADIST}:5--12, \textit{the outer loop}) with $m$ expanding attempts to find valid paths for the agents from node $v$. \added{Making repeated planning attempts before backtracking enable paths to resume expansion from stalled positions and avoid local conflicts; our experiments (Figure~\ref{fig:peNr-varying}) show that even just a few replanning attempts greatly helps to improve success rates with negligible additional computation.}
\added{This outer} loop begins by setting the variable $\Var{allRG}$ (which stands for \textit{all reach goal}) to $\True$ (Alg.~\ref{alg:MADIST}:6), \added{then it goes into the \textit{inner loop} (Alg.~\ref{alg:MADIST}:7--11)}.

The \added{inner} \emph{for} loop plans for each agent by calling the module \saplannerdl{} then \added{inserting each returned path into the trajectory array $\Traj^s_{p_i}$ (Alg.~\ref{alg:MADIST}:9)}. If \added{the returned path} $\Traj^s_{p_i}$ is \textit{reach-goal},
its validity is checked by the function $\Function{ModifyIfOverlap}()$. 
The path is invalid if the \added{agent stops at its goal} at time $t$ and blocks \added{the future movement (after $t$)} of an already-planned agent (according to planning order $p$), a situation called \textit{collision-at-goal} shown in Figure~\ref{fig:fw}.2--3.
If this situation occurs, the function determines the collision time $t_c$, trims the already-planned paths (those which extend beyond $t_c$) to $t_c$; then sets $\Var{allRG}$ to $\False$ (Alg.~\ref{alg:MADIST}:10). 
If the path is not reach-goal, $\Var{allRG}$ is set to $\False$~(Alg.\ref{alg:MADIST}:11). 

After \added{the planning for all agents has completed, the tree $\Tree_{tc}$ is expanded from $v$ by the function $\Function{ExpandTCTree()}$} using the agent's paths $\Traj^s$ (Alg.~\ref{alg:MADIST}:13).
If all agents reach the goals ($\Var{allRG}$ is still $\True$), the \emph{while} loop is broken (Alg.~\ref{alg:MADIST}:14), then the planner returns paths for all agents from $\Tree_{tc}$ (Alg.~\ref{alg:MADIST}:15--16).

\begin{algorithm}[t]
  \caption{Low Level: Single-Agent Pathfinding with Dynamic Leading (\saplannerdl)}
  \label{alg:sapf}
  \begin{algorithmic}
  \STATE \text{INPUT}: $\rho$: \text{agent's ID}, $g_\rho$: goal, $\mathcal{G}$: init. graph, $\Traj^s$: \text{previous planned paths}, 
  $t_\Var{sa}$: \text{max runtime}, $\mathcal{H}_\rho$: \text{shortest path heuristic}
  \STATE \text{OUTPUT}: a path $\Traj_\rho$
  \end{algorithmic}
  \begin{algorithmic}[1]
  \STATE $\mathcal{G}^s = \Function{ModifyGraph}( \mathcal{G}, g_\rho, \Traj^s_\rho.\Var{end})$;
  $v_g\!\gets\! \Function{GetNode}(g_\rho,\mathcal{G}^s)$; 
  $v_s\!\gets\!\Function{GetNode}(\Traj^s_\rho.\Var{end}, \mathcal{G}^s)$\;
  \STATE $\Var{openList}.\Function{Add}(v_s)$; $\Var{closedSet} \gets \{\}$; $v_\Var{best} \gets v_s$\;
  
  \WHILE{$\Var{openList}$ \textbf{not empty} \textbf{and} $\Function{time}() < \Var{t}_\Var{sa}$}
      \STATE $v \leftarrow \Var{openList}.\Function{Pop}()$; $\Var{closedSet}.\Function{Add}(v)$\;
      \STATE \textbf{if} $v = v_g$ \textbf{then return} $\Function{getpath}(v_s, v,\mathcal{G}^s)$ 
      \FOR{$u$ \textbf{in} $\Function{GetNeighbors}(v)$}
          \IF{$u  \textbf{ is not in }  \Var{closedSet}$}
              \STATE $u.\Var{parent} \gets v$; $d_{uv} \gets \Function{Distance}(v, u)$; $u.g \leftarrow v.g$ + $d_{uv}$\;
              \STATE $u.t \gets v.t$ + $d_{uv}/\text{v}_c$; $u.h \gets \Function{GetHeuristic}(u, \mathcal{H}_\rho)$; $u.f=u.g+u.h$\;
              \IF{$\Function{IsActionValid}$($\rho, \Traj^s, u, v, r_c)$}
                  \STATE $\Var{openList}.\Function{Add}(u)$\;
                  \STATE \textbf{if} $u = v_g$ \textbf{then return} $\Function{GetPath}(v_s,u,\mathcal{G}^s)$\;
                  \STATE \textbf{if} $v_\Var{best}.f > u.f$ \textbf{then} 
                  {$v_\Var{best} \gets u$}\;
              \ENDIF
          \ELSIF{$v.g +d_{uv} < u.g$}
              \STATE $u_1 \gets \Function{Copy}(u)$; $u_1.t \gets v.t+d_{uv}/\text{v}_c$\;
              \IF{$\Function{IsActionValid}$($\rho, \Traj^s, u_1, v, r_c$)}
                  \STATE $u.\Var{parent} \gets v$; $u.t \leftarrow v.t+d_{uv}/\text{v}_c$;
                  $u.g \gets v.g+d_{uv}$; $u.f=u.g+u.h$\;
                  \STATE \textbf{if} $v_\Var{best}.f > u.f$ \textbf{then} {$v_\Var{best} \gets u$}\;
              \ENDIF
          \ENDIF
      \ENDFOR
  \ENDWHILE
  \STATE \textbf{return} $\Function{GetPath}(v_s, v_\Var{best},\mathcal{G}^s)$\;
  \end{algorithmic}
\end{algorithm}

\subsection{Low Level: Single-Agent Pathfinding with Dynamic Leading}
\label{sec:SAPF}
\added{Under the continuous-time \gls{ACOMM} constraint, \glsentryfull{SAPFDL} planner---an A$^*$-based search algorithm---needs several key features. First, a timestep must be computed for each node added into the $\Var{openList}$. The input also include the start and goal positions for each agent.}
Second, the \textsc{IsActionValid}() function (Alg.~\ref{alg:isActionValid}) verifies whether the current action is in collision-free and satisfies \acomm{} constraint \added{in continuous time, i.e., along the entirety of each trajectory segment}. Our proposed \textit{dynamic leading} technique is integrated within this function to dynamically reassign the leader role to the planning agent if it is the most progressive one.
Third, to improve the search efficiency in complex obstacle structures and cluttered spaces, we employ shortest paths to the goals from all sub-divisions as a heuristic, obtained via the \textsc{GetHeuristic}() function.

The inputs to this low-level planner include the agent's \textsc{id} $\rho$, goal position $g_\rho$, a graph $\mathcal{G}$ representing the environment, the planned paths of other agents $\Traj^s$, and the shortest path heuristic $\mathcal{H}_\rho$.
\added{The input graph $\mathcal{G}$ must also include the agent's current position $\Traj^s_\rho.\Var{end}$ and its goal $g_\rho$. The modification process involves: (1) identifying the two subdivisions that contain the current position and goal, (2) locating the corresponding vertices in graph $\mathcal{G}$, and (3) updating the vertices' positions---originally the subdivision centroids---to the $\Traj^s_\rho.\Var{end}$ and the goal $g_\rho$, respectively. All these steps are implemented by the $\Function{ModifyGraph}$() function (Alg.~\ref{alg:sapf}.1)}.
\added{The variables $v_s$ and $v_g$ represent the current and goal nodes, respectively (Alg.~\ref{alg:sapf}:1).
The node $v_s$ is then added into $\Var{openList}$, a priority queue that sorts elements based on their cost-to-goal. 
The closedSet and the variable representing the closest-to-goal node, $v_\text{best}$, are also initialized (Alg.~\ref{alg:sapf}:2).}

\added{The main search loop (Alg.~\ref{alg:sapf}:3--18) operates as follows. The node with lowest cost is popped out from the $\Var{openList}$ and inserted into the $\Var{closedSet}$ (Alg.~\ref{alg:sapf}:4). If this node is the goal, search terminates successfully (Alg.~\ref{alg:sapf}:5).}
\added{Otherwise, all of its neighbors are iterated through (Alg.~\ref{alg:sapf}:6--18) to identify valid nodes, compute or update their costs, parent nodes, and timestep, then insert them into $\Var{openList}$. For each node $u$ not in the $\Var{closedSet}$, we check its validity by determining if the \textit{move} action from parent $v$ satisfies both the collision-free and the \acomm{} constraints, as evaluated by the $\Function{IsActionValid}$() function (Alg.~\ref{alg:sapf}:10). If $u$ is already in the $\Var{closedSet}$, its estimated cost is updated only if the new cost is smaller than the existing one (Alg.~\ref{alg:sapf}:14-17).}
\added{The \Function{GetNeighbors}() returns both physical neighbors and, if a \textit{wait} action is included in the set of available actions, the current node. When the \textit{wait} action is selected, the waiting time is one second, with a corresponding cost v$_c$ (Alg.~\ref{alg:sapf}:6,8)}

\begin{algorithm}[t]
  \caption{$\Function{IsActionValid}(\rho, \Traj^s, u, v, r_c$) (\textit{Dynamic leading employed within this function}) }
  \label{alg:isActionValid}
  \begin{algorithmic}[1]
  \STATE $\Var{lead} \leftarrow \True$; $\Var{comm} \leftarrow \False$\;
      \FOR{ $1 \leq i \leq |\Traj^s |$ \RightComment{dynamic leading (line:2-3)}}
          \STATE \textbf{if} $u.t \leq \Traj^s_i.\text{maxtime}$ \textbf{then} $\Var{lead} \gets \False$; \RightComment{is this agent most progressive?}
      \ENDFOR
      \STATE \textbf{if } \text{lead} \textbf{then} $\text{lead} \gets \Function{IsCOMMAtGoal}(u, \Traj^s, \rho)$\;
      \FOR{$1 \leq i \leq |\Traj^s|$}
          \STATE \textbf{if} $\Function{IsCollision}(u, v, \Traj^s_i, \rho)$ \textbf{then return} $\False$\;
          \STATE \textbf{if} $\Function{IsCOMMS}(u, v, \Traj^s_i, \rho, r_c)$ \textbf{then} $\Var{comm} \leftarrow \True$\;
      \ENDFOR
      \STATE \textbf{return} {$(\Var{lead} \ \| \ \Var{comm})$ } \RightComment{leader moves without \gls{ACOMM} constraint}
  \end{algorithmic}
\end{algorithm}

\paragraph{\textbf{\textsc{IsActionValid}}() Function}
This function (Alg.~\ref{alg:isActionValid}) checks the validity of action $\textit{Move}(v,u)$ for an agent $\rho$ starting at time $v.t$. To be valid, the action must maintain \gls{ACOMM} constraint and collision free as the agent implements the action. The \gls{ACOMM} constraint is satisfied by one of two conditions: having a communication link with at least one neighbor or being a leader.

The \textit{dynamic leading} technique allows the agent $\rho$ to take the leader role if its action goes furthest in time (Alg.~\ref{alg:isActionValid}:2--3). 
The leader is exempt from the \acomm{} constraint.
The leader agent is determined by comparing the timestep of current action to the max-time of all the paths (Alg.~\ref{alg:isActionValid}:2--3).
For agent-collision detection, $\Function{IsCollision}()$ (Alg.~\ref{alg:isActionValid}:6) checks if any segmented point along the action \textit{Move}($v,u$) collides with the planned paths in $\Traj^s$ (Alg.~\ref{alg:isActionValid}:5) at the corresponding timesteps. 
The action is also checked by function $\Function{ISCOMMS}()$ (Alg.~\ref{alg:isActionValid}:7) for the \acomm{} constraint, that determines whether agent $\rho$ can communicate with a neighbor during the move. 

As deploying \textit{dynamic leading}, a situation called \textit{out-of-communication-at-goal} can occur, which is shown in Figure~\ref{fig:outComAtGoal}a. 
At $t_i$ agent $a_3$ is the leader, and reaches its goal in two steps ($t_{i+1}$ and $t_{i+2}$). $a_2$ follows $a_3$ and also stops at its goal by $t_{i+2}$.
Agent $a_1$ moves along $a_3$ until $t_{i+2}$, then becomes the leader, planning without \acomm{} constraints. Agents $a_2$ and $a_3$ reaches the goals and stop there, thus the \gls{TCOMM} constraint breaks at $t_{i+6}$. 
To resolve this issue, we propose the function $\Function{IsCommAtGoal}$().

\paragraph{\textbf{\textsc{IsCommAtGoal}}() Function}
The function $\Function{IsCommAtGoal}()$ guides $a_1$ to remain in communication as shown in Figure~\ref{fig:outComAtGoal}b. It is triggered when the leader is changed (Alg.~\ref{alg:isActionValid}.4) to check if any of the new leader's neighbors have reached their destination. If so, the leader's action $\textit{Move}(v,u)$ must meet the \acomm{} constraint with an at-goal agent. If it cannot, its leader status is revoked.

\subsection{Expand Team Communication Tree}
\label{sec:TCT_alg}
Function $\Function{ExpandTCTree}()$ (Alg.~\ref{alg:ExpandTCTree}) expand the \gls{TCT} \tctree{} by integrating the agents' paths $\Traj^s$.
Each node on $\Tree_{tc}$ has an associated timestep, $f$-value, and positions are interpolated to make them match for agent paths.
\added{A new node must satisfy the \tcomm{} constraint (function \textsc{IsTCOMM}()) to be added to the tree}.
The \emph{for} loop (Alg.~\ref{alg:ExpandTCTree}:2--3) collects all timesteps of the waypoints on the agent paths and inserts into a priority queue $\Var{timeList}$. 

\begin{algorithm}[t]
  \caption{$\Function{ExpandTCTree}$(\tctree, $\Traj^s, v$) }
  \label{alg:ExpandTCTree}
  \begin{algorithmic}[1]
  \STATE $v_p \gets v$; $\Var{timeList} \gets \{ \} $ \RightComment{priority queue}\;
  \FOR{$0 \leq $ i $ < | \Traj^s|$ }
      \STATE \textbf{for} $0 \leq $ j $ < | \Traj^s_i.\Var{times} |$ \textbf{do} \Var{timeList}.$\Function{Insert}(\Traj^s_i.\Var{times} [j])$ \;
  \ENDFOR
  \FOR {$t$ \textbf{in} \Var{timeList} }
      \STATE $v_n \leftarrow \Function{NewNode}()$; $v_n(\Var{parent}, g,t) \leftarrow (v_p,v_p.g,t)$\;
      \FOR{$0 \leq j < | \Traj^s|$} 
          \STATE \textbf{if } $t > \Traj^s_j.\Var{end}.t$ \textbf{then}\;
          \STATE \quad \textbf{if } $\Function{FoundGoal}(\Traj^s_j)$  \textbf{then} $\Var{pos} \leftarrow \Traj_j.\Var{end.pos}$\;
          \STATE \quad \textbf{else return}\;
          \STATE \textbf{else} $\Var{pos} \leftarrow \Function{GetPosAtTime}(\Traj^s_j, t)$ \;
          \STATE $v_n.\Var{state}.\Function{Insert}(\Var{pos}) $; $d = \Function{Dist}(v_p.\Var{state}_j, v_n.\Var{state}_j)$;
          \STATE $v_n.g$ += $d$; $\Traj^s_j$.\Var{costogoal} -= $d$;
          $v_n.h$ += $\Traj^s_j$.\Var{costogoal}\;
      \ENDFOR
      \STATE \textbf{if } \Function{IsTCOMM}($v_n$) \textbf{then} $v_n.f \gets \alpha v_n.g$ + $(1-\alpha)v_n.h$;
      $\Tree_\Var{tc}.\Function{Append}(v_n)$; $v_p \gets v_n$\;
      \STATE \textbf{else return} 
  \ENDFOR
  \end{algorithmic}
\end{algorithm}

\begin{figure}[b]
  \centering
  \includegraphics[width=0.6\columnwidth]{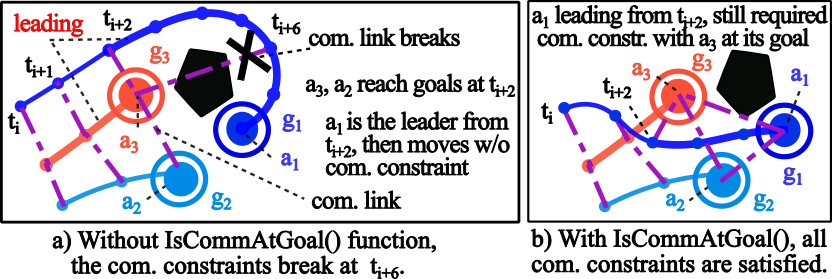}
  \caption{\textbf{Out-of-communication-at-goal situation (a) and how 
  \Function{IsCommAtGoal}() works (b).} In both situations, the planning order 
  is ($a_3, a_2, a_1$) and the leader is changed at $t_{i+2}$.}
  \Description{Out-of-communication-at-goal situation occurs as the leader agent can plan without \gls{ACOMM}, while other agents are already at their goals and not follow the leader, breaking the team communication channel. Our function \Function{IsCommAtGoal}() prevents it by revoking the leader status of current leader if its neighbors have reached the goals already.}
  \label{fig:outComAtGoal}
\end{figure}

The second \emph{for} loop (Alg.~\ref{alg:ExpandTCTree}:4--14) goes through all 
elements in $\Var{timeList}$ to create new nodes and adds valid nodes into the tree $\Tree_\text{tc}$.
Each new node $v_n$ inherits the travel cost from its parent with timestep $t$ (Alg.~\ref{alg:ExpandTCTree}:5). 
We then go to each agent's path, interpolate to recover its position at time 
$t$ (Alg.~\ref{alg:ExpandTCTree}:7--12), and insert it into the state of node $v_n$.
If $t$ is larger than the max-timestep of the path $\Traj^s_j$ (Alg.~\ref{alg:ExpandTCTree}:7) 
and the path reaches the goal, the final position of $\Traj^s_j$ is returned (Alg.~\ref{alg:ExpandTCTree}:8). 
If the position is not the goal, the tree's expansion stops (Alg.~\ref{alg:ExpandTCTree}:9).

If $t$ is smaller or equal to the last timestep $\Traj^s_j$, the agent's position at $t$ is interpolated by function $\Function{GetPosAtTime}()$ (Alg.~\ref{alg:ExpandTCTree}:10). 
The lines in Alg.~\ref{alg:ExpandTCTree}:11--12 add the agent's positions, cost, and heuristic of each agent into the node state.
The new node is then verified the \gls{TCOMM} constraint (preventing the scenario in Figure~\ref{fig:outComAtGoal}a) before adding to the tree \tctree (Alg.~\ref{alg:ExpandTCTree}:13). If the \gls{TCOMM} constraint is not satisfied, the tree expansion is terminated (Alg.~\ref{alg:ExpandTCTree}:14). 

\subsection{Algorithm Complexity Analysis}
\label{sec:alg_analysis}
The runtime of the High-Level module (Alg.~\ref{alg:MADIST}) is $O(k)$ where $k$ is the number of \gls{TCT} expansions. For each expansion, the low level \saplannerdl{} (Alg.~\ref{alg:sapf}), based on \textsc{a}* search, is called $m \cdot n$ times, where $m$ is the max number of single-agent path expanding attempts per agent (Alg.~\ref{alg:MADIST}:5) and $n$ is the number of agents.
The \textsc{a}* search with the shortest path heuristic has a complexity of $O(|E| \cdot \text{log} (|E|))$, where $|E|$ is the number of edges, modified by a valid action check whose runtime, dependent on collision checking and the communication model with worst case is $O(n^2)$ for all pairwise constraint checks. Thus, the overall complexity is approximately $O(k\cdot m\cdot n\cdot |E| \text{log} |E|\cdot n^2) = O(k \cdot m \cdot n^3\cdot |E| \text{log} |E|)$.
The performance of \planner{} planner depends on the selection of \gls{TCT}'s nodes
and whether the \textit{out-of-communication-at-goal} situation occurs.
The current heuristic---sum of agents' shortest paths---is efficient to identify nodes with the shortest travel distance to goals but fails to consider the communication constraints, making it ineffective at selecting nodes to prevent the \textit{out-of-communication-at-goal} issue.
In future research, it may be possible to develop a more informative, if more computationally demanding, heuristic to select nodes from \gls{TCT} that incorporates communication costs and avoids the \textit{out-of-communication-at-goal} situation.

\section{Experiments and Results}\label{sec:expres}
Experiments are conducted on five obstacle-rich environments (Figure~\ref{fig:result-agentNr}). 
The planning performance is measured by three metrics: (1) success-rate, (2) runtime, and (3) per-agent travel distance. 
Our planner, \planner, is evaluated with five baselines: a centralized approach with composite states, a platooning leader-follower approach, modified versions for the \malcr{} problem of three planners: \gls{PIBT}, \textsc{od-id}, and \gls{PBS}. All agents have 8 actions: to move in the four cardinal directions and in their diagonal directions. 
Although the actions are discrete, the agent's positions are continuous, with random starts, goals, and the interpolated positions from agent paths. 
\added{We also compare the performance of our \planner{} when a \textit{wait} action is included (referred to as \planner{}-wait) and without it (\planner{})}. 
We evaluate the planner's performance on two types of communication constraints: \glsentryfull{LCR} and \glsentryfull{LOS}; and the results are described in Section~\ref{sec:dist_results} and Section~\ref{sec:los_results}, respectively. We also test the planners versus runtime, goal configurations, and difficulty of environments.

\subsection{Experimental Setup}

\subsubsection{Baselines} \label{exp:baselines}

We have implemented five baselines: \gls{COMP}, \gls{PLF}, \gls{ODIDCOMM}, \gls{PIBTCOMM}, and \gls{PBSCOMM}.
\gls{COMP} is selected because it theoretically can solve the \gls{MAT3C} problem. 
\gls{PLF} is chosen because it is the state-of-the-art algorithm to handle \gls{MAT3C} problem in the leader-follower approach.
Meanwhile, \textsc{od-id}, \gls{PBS}, and \gls{PIBT} are state-of-the-art algorithms in standard \gls{MAPF} instances, which were shown efficient to plan for a large number of agents. 
Both \gls{PBS} and \gls{PIBT} belong to decoupled approach with different single-agent path expansion strategies: single-expansion and single-action in each planning iteration. 
We select \textsc{od-id} to represent for the composite state approach over \textsc{m}* because agents in \textsc{m}* select a single action per planning iteration based on their own interests (similar to \textsc{pibt}), which can cause follower agents to fail in maintaining communication constraints with the leader agent.
Difference to general \gls{MAPF} instances, in \gls{MAT3C} problem, actions are only valid if they satisfy the \acomm. Consequently, planners must establish planning orders, ensuring that later-planned agents maintain the \acomm{} constraint with previously planned agents, and if one agent is replanned, all agents subsequent to it in the planning order must also be replanned. Thus, these planners are algorithmically modified to fit \gls{TCOMM} requirement, and referred as \gls{ODIDCOMM}, \gls{PBSCOMM}, \gls{PIBTCOMM}, respectively.

\begin{itemize}
    \item \Gls{COMP} grows a multi-agent tree with composite states of all agent's positions via \textsc{a}$^*$ with a heuristic that sums shortest-to-goal paths from each agent.
    \item \Gls{PLF} also builds a multi-agent tree and uses priority planning to grow the tree. At the root of tree expansion, the planning order is shuffled and so a random leader is chosen; planning order is then fixed for downstream nodes.
    The followers plan to follow another agent if its preceding agent reached its goal.
    If the current expansion fails after some iterations, the planning starts again at the root with a new random leader and a corresponding planning order.
    \item \textsc{Od-id} initially computes individual paths for each agent, then uses the \textsc{id} technique to identify conflicts and replan one agent's path to avoid collisions while keeping other paths fixed. 
    If no valid path is found, the planner merges the conflicting agents and plan on their composite states with the \textsc{od} technique.
    Different to \textsc{od-id}, \gls{ODIDCOMM} needs to select the first agent randomly to plan with \acomm-free, requiring subsequent follower agent's actions to satisfy \gls{ACOMM} constraint relative to the already planned paths. As planning on composite states, composite actions must satisfy the \gls{TCOMM} constraint and collision free.
    \item \Gls{PIBT} is a state-of-the-art \gls{MAPF} algorithm that uses the \textit{priority inheritance} scheme to move agents by single actions respect to their own interest in each iteration. To work on the \malcr{} problem, we modified \gls{PIBT} by adding \acomm{} into the action selection of the agents except the first planning agent. We refer to this planner as \gls{PIBTCOMM}.
    \item \Gls{PBS} uses a priority tree to resolve conflicts among agents. To handle the communication constraint in the \malcr{} problem, we develop a modified version of \gls{PBS}, named \gls{PBSCOMM}. In \malcr{}, neighbors at starting positions are possible not neighbors at goals, thus \gls{PBSCOMM} needs multiple iterations to find the feasible planning order for the entire agent team with \acomm{} at the root node. In many cases, the planner fails because no feasible planning order exists or planning runs out of time with large number of agents. The second change is to add a planning order due to the communication constraint. That leads to a modification of replanning implementation: if an agent with priority $p_i$ is replanned, all agents with lower priority than the agent are also recomputed due to the communication dependence. These two changes, necessary to apply \gls{PBS} to the \malcr{} problem, significantly increases runtime.
\end{itemize}

\subsubsection{Environments and Instances} 
\begin{figure}[t]
  \centering
  \includegraphics[width=1.0\textwidth]{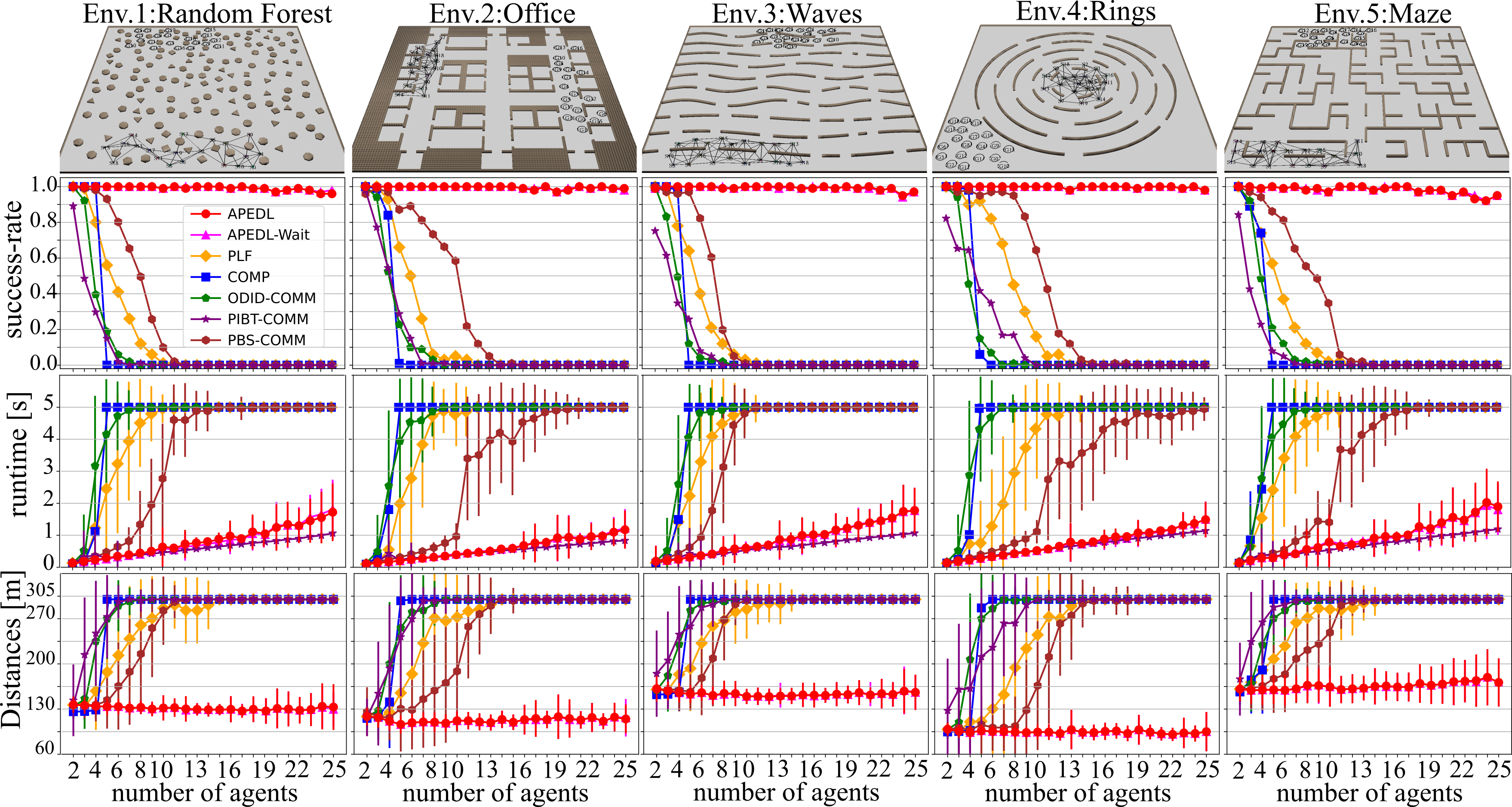}
  \caption{\textbf{Success-rate, Runtime, and Travel Distances of our planners: \planner{} and \planner-\textsc{e} with \gls{PLF}, \gls{COMP}, \gls{ODIDCOMM}, \gls{PIBTCOMM}, and \gls{PBSCOMM} on 5 environment types under \gls{LCR} constraint}. 
  The comm. range is 15 m and max-runtime is 5 s. The runtime and travel distances are shown by means and standard deviations.}
  \Description{The experiment's results with success-rate, runtime, and distances are respect to number of agents. Our planer \planner{} can compute paths up to 25 agents in all five environments within 5 seconds while other baselines fail.}
  \label{fig:result-agentNr}
\end{figure}

We evaluate \planner{} in five obstacle-rich environment types: Random Forest, Office, Waves, Rings, and Maze (Figure~\ref{fig:result-agentNr}). All environments are square shapes of size $114\text{m} \times 114\text{m}$ , which are divided into sub-divisions with resolution of $1\text{m} \times 1\text{m}$. Without considering the obstacles, the sub-divisions create a graph with approximate 12700 vertices and 51000 edges. 
On each environment type, there are one hundred maps generated with random locations of obstacles.

For each number of agents, one instance is generated on each environment map. 
So, there are total of 12000 instances for the experiments. 
The start and goal configurations are chosen randomly, subject to the \tcomm{} constraint (satisfying both \gls{LCR} and \gls{LOS} constraints), on alternate sides of the maps (except the ring environment), distributed with a rectangular area. 
Communication between two agents depends on the communication model. For \gls{LCR}, agents can communicate if the distance between them is within 15 m. For \gls{LOS}, communication is established if the agents have an unobstructed view of each other.

\paragraph{Env. Type 1: Random Forest} Obstacles with random shapes and sizes are distributed randomly occupying 10\% of the environment's area.

\paragraph{Env. Type 2: Office} The environments consist multiple rooms and hallways. Each room has a fixed width and varying length from 9--13 m. There are three long hallways along the building, 2--3 short hallways connecting them. Hallway width varies from 7--9 m.   

\paragraph{Env. Type 3: Waves} The environments have wave-like obstacles which are separated at regular distances. Gaps are placed in each wave with random widths. The number of waves is set to $10$.

\paragraph{Env. Type 4: Rings} The environments are featured by concentric rings with six breaks of random widths within 6--8 m. The separation between the rings is set to 8 m. For each instance, the starts are randomly placed at the center, and the goals are on one of four corners of the maps.

\paragraph{Env. Type 5: Maze} The environments consist of mazes generated by Kruskal's algorithm with size of $14 \times 14$. To ease in generation of the starting and goal configurations, boundary walls connecting to the top-most and bottom-most rows are removed.

\subsubsection{Measuring Performance} 
We evaluate \planner{} and \planner-\textsc{e} against \gls{COMP}, \gls{PLF}, \gls{ODIDCOMM}, \gls{PIBTCOMM}, \gls{PBSCOMM} baselines on 100 instances for each environment type and number of agents, measuring success-rate, average runtime, and average distance traveled per agent. 
Planning is considered successful if all agents reach their goals within 5 s of runtime. Runs that exceed that planning ceiling are assigned a travel distance of 300 m.
The framework is evaluated with variations over number of agents, goal configurations, runtime, and environment difficulty levels. 

\subsubsection{Computing Resources} 
The experiments ran on \textsc{hopper}, a computing cluster provided by \textsc{gmu}'s Office of Research Computing. Each planning instance was run single threaded, yet experiments were run in parallel across 48 cores on a 2.40GHz processor. Our code was developed in C++ and compiled with g++-9.3.0.

\subsection{Results Under Limited Communication Range Constraint}
\label{sec:dist_results}
In these experiments, we deployed \gls{LCR} as the communication constraint on the agent team. Communication between two agents is established if their distance is smaller than a threshold $d_c$ ($d_c$=15 m in the experiments). We evaluate the planner's robustness versus the number of agents (Section~\ref{sec:dist_nr}), runtime (Section~\ref{sec:dist_rt}), environment difficulty (Section~\ref{sec:dist_envdif}), and goal configurations (Section~\ref{sec:dist_gconf}).

\subsubsection{Results when varying the agent number} 
\label{sec:dist_nr}
We evaluated the planners with 2--25 agents, allocating 5 seconds of runtime, and the results are shown in Figure~\ref{fig:result-agentNr}. 
The \planner{} planner performs well up to 25 agents exceeding 90\% success-rate in all environments. The \gls{APG} technique iteratively grows and refines the agent paths, enabling the \planner{} planner to effectively handle arbitrary start and goal configurations. The \gls{DL} technique allows leader to be reassigned in the same planning order during each planning iteration, making followers resume the leader roles and progress to their goals as the current leader stalls.
\added{We additionally show that the performance of \planner{} with the \textit{wait} action (\planner-wait) is similar, suggesting that there is little benefit in our approach to having later agents pause during their motion.}

All baseline planners struggle as the number of agents increases. \gls{PLF} manages up to 5 agents in Rings Env. and only 3--4 agents in other environments with 90\% success-rate. 
The heuristic of shortest-to-goal path sum guides \gls{COMP} well to handle up 4 agents achieving 90\% success-rate in Random Forest, Rings, and Waves Env. However, with greater than 4 agents, the high dimensionality of the search space means that planning cannot reach the goal in the allotted time and success rate quickly declines.

The \gls{PBSCOMM} baseline outperforms the other baselines in the \gls{MAT3C} problem. Its priority tree effectively determines feasible planning orders for conflicting agents and efficiently identifies the best order. However, relying solely on planning order is insufficient to fully address the \malcr{} problem, as neighbors of agents change as navigating from start and goal configurations, often rendering feasible planning orders unattainable. Consequently, the performance of \gls{PBSCOMM} decreases from 5--7 agents (with 90 \% success-rate) in Waves, Random Forest, and Rings environments and with as few as 3--4 agents in the more constrained Office and Maze environments.

\Gls{PIBTCOMM} exhibits the poorest performance due to its myopic action selection, which prioritizes actions based on a shortest path heuristic in each planning iteration. This approach frequently leads to situations where follower agents cannot select actions to maintain the communication channel with the leader in subsequent iterations. As a result, \gls{PIBTCOMM} failed even with as few as three agents across all tested environments.

\begin{figure}[t]
  \centering
  \includegraphics[scale=0.40]{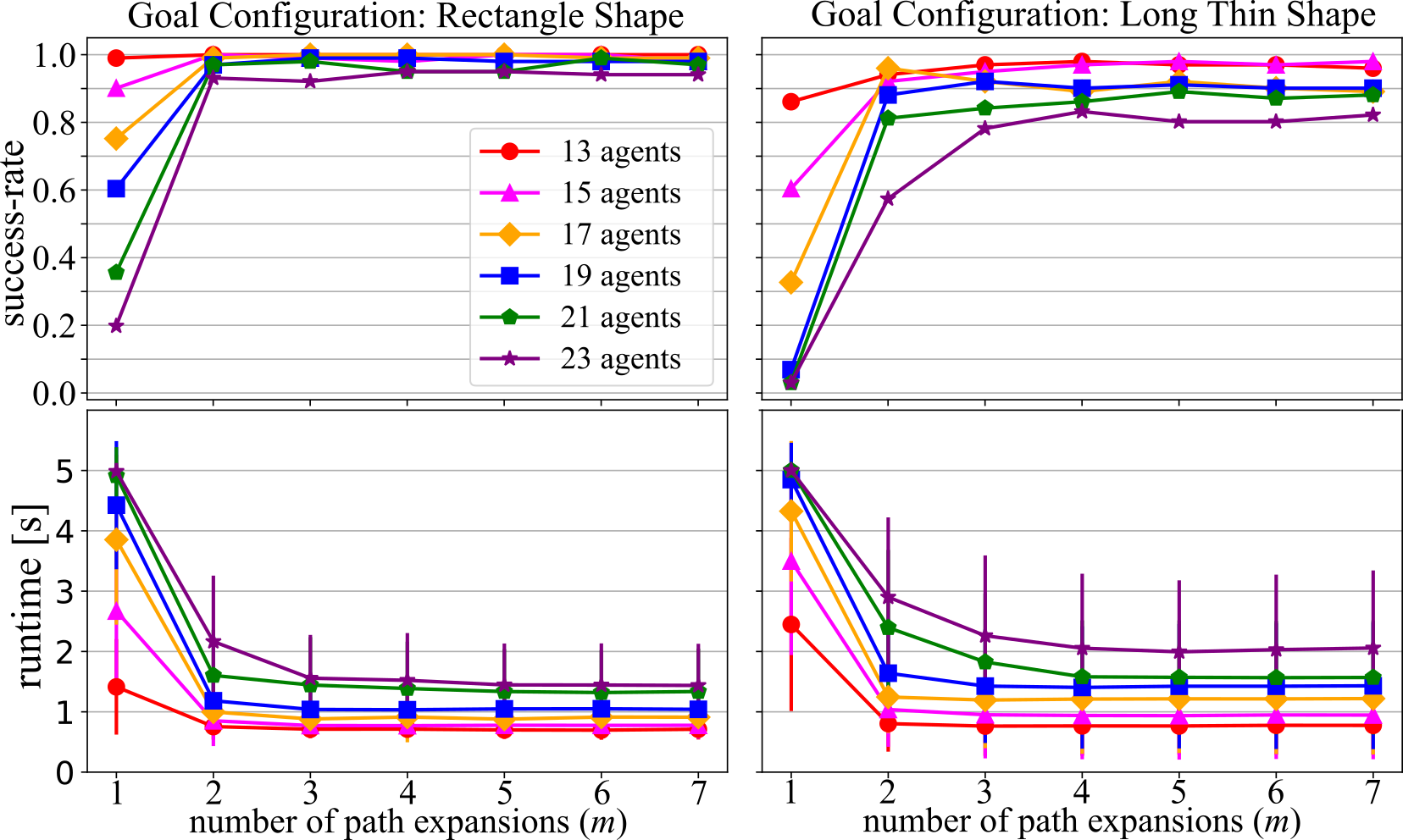}
  \caption{\textbf{Performance of \planner{} as varying the number of path expansions in Maze environments (Env.5)}. One-time path expansion ($m=1$) fails as the number of agents increase. Adaptive path Expansion (\textsc{ape}) ($m>1$) can compute paths efficiently even with Goal Configuration in Long-Thin Shape.}
  \Description{These results emphasize the effectiveness of our proposed technique, \textsc{ape} ($m > 1$). In contrast, the one-time path expansion approach ($m = 1$) struggles to find feasible paths as the number of agents increases or when the goal configurations become more challenging.}
  \label{fig:peNr-varying}
\end{figure}

The \gls{ODIDCOMM} algorithm first plans paths for individual agents, then addresses collisions by prioritized planning, and if necessary, employs the \textsc{od} technique for composite states. In the \malcr{} problem, where agents are closely positioned and collision are frequent, \gls{ODIDCOMM} allocates part of its runtime to single-agent planning and the remainder to composite-state planning. In contrast, \gls{COMP} planner focuses entirely on composite states, leading to better performance in high-collision scenarios. Consequently, \gls{ODIDCOMM} generally underperforms compared to \gls{COMP} planner in the \gls{MAT3C} problem due to this runtime distribution. However, in simpler cases with up to five agents, \gls{ODIDCOMM} successfully completes planning, while the \gls{COMP} planner completely fails due to the exponential growth of its composite state space.

In addition to average success rate, we also report the mean and standard deviation for runtime and travel distance in Figure~\ref{fig:result-agentNr}. If a planner fails, the runtime and traveled distance penalties are applied to the results. The runtime and distance results aligned with the success-rate results across all planners except \gls{PIBTCOMM} planner. The \gls{PIBTCOMM} planner exhibits low runtime due to early termination when follower agents fail to maintain communication with the leader, limiting its effectiveness in the \gls{MAT3C} problem. In contrast, as the number of agents increases, the \gls{PBSCOMM} planner allocates significant runtime to computing a feasible \acomm{} planning order from all possible agent sequence combinations, hindering its efficiency in resolving the \malcr{} problem.  

\subsubsection{\added{Results when varying number of path expansions}}

\added{This experiment evaluated the impact of number of path expansions $m$ to the performance of \planner{} on two goal configurations within the Maze environment: rectangle shape and long-thin shape. 
The results include the success-rate and runtime and are illustrated in Figure~\ref{fig:peNr-varying}.}
\added{When the number of path expansion attempts is 1 ($m=1$), the \planner{} only computes paths from each node of \tctree{} only once (similar to \gls{PBS}, \gls{OTIMAPP}). With more than 13 agents or in challenging goal configurations, making only a single attempt considerably limits the success-rate. By contrast, increasing $m$ improves performance without increasing planning cost and in fact decreasing plan cost since suitable paths are found more quickly. In all other experiments, we use a value of $m=5$.}

\subsubsection{Results when varying environment difficulty}
\label{sec:dist_envdif}
\begin{figure}[t]
\setlength{\tabcolsep}{1pt}
\centering
\includegraphics[width=0.8\textwidth]{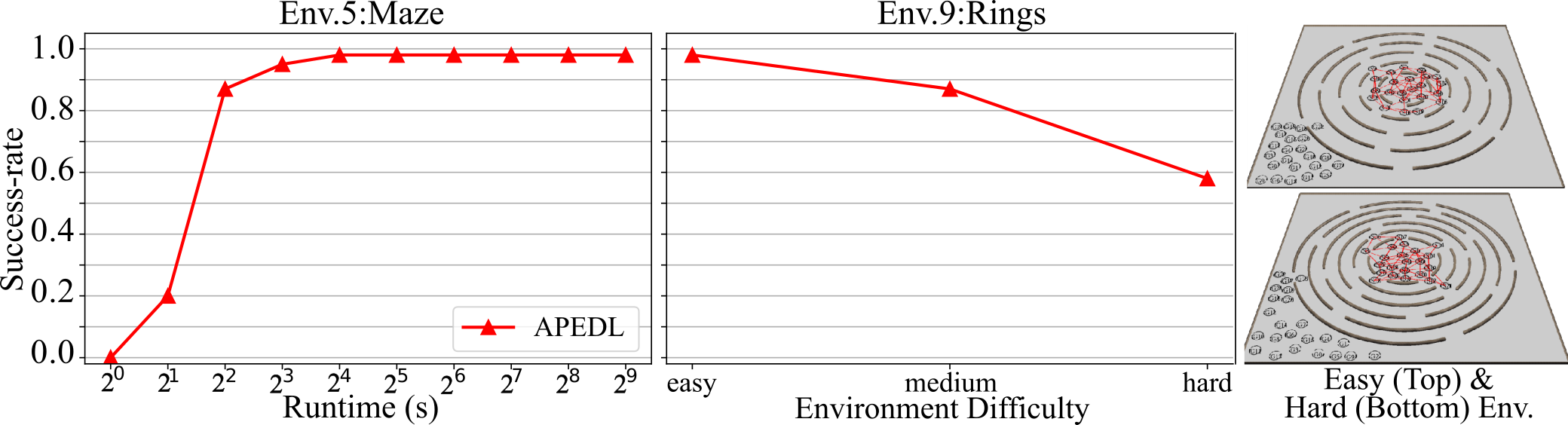}
\begin{tabular}{l c}
    { \footnotesize{a) Success-rate vs Runtime}} \qquad \qquad \qquad \qquad &
    {\footnotesize{b) Success-rate vs Env. Difficulty}} 
\end{tabular}
\caption{\textbf{Performance of \planner{} as varying the runtime (a) and environment difficulty (b) with 23 agents}.}
\Description{We evaluate our planner (\textsc{apedl}) respect to runtime (from 1 to 64 seconds) and Environment Difficulty (Easy, Medium, and Hard). The planner performance is better as increasing the runtime or reducing the challenges of obstacle-structures.}
\label{fig:rt_envhardness}
\end{figure}

To demonstrate the robustness of our framework, we run \planner{} on variations of the Rings environment type with three difficulty levels: \textit{easy}, \textit{medium}, and \textit{hard}. The difficulty is controlled by the number of concentric rings, distorderance between consecutive rings, and number of breaks/gates on the rings. 
The features controlling difficulty are shown in  Table~\ref{tab:hardnessEnvPara}. An easy and a hard environment are illustrated in Figure~\ref{fig:rt_envhardness}.b. 
The performances on three environment types with 23 agents and 5 s runtime are shown in Figure~\ref{fig:rt_envhardness}.b with success-rate decreasing from easy to hard level. With medium or hard levels, \planner{} planner needs more than 5 s runtime to find valid paths for the agents.

\begin{table}[h]
\caption{Environment Difficulty Features On Rings Environments}
\label{tab:hardnessEnvPara}
\centering
  \footnotesize
  \begin{tabular}{c|ccc}
      \toprule
      Difficulty Level & Ring Count & Ring Distance & Break Count\\
       \midrule
       Easy & 4--5& 8.0 m & 6--7 \\
       Medium & 5& 7.0 m & 5--6 \\
       Hard & 6 & 5.5 m & 4--5\\
       \bottomrule
  \end{tabular}
\end{table}

\subsubsection{Results when varying runtime}
\label{sec:dist_rt}
In this experiment, we ran the planners with various runtime from 1 s to 512 s. We selected the Maze Env., the most challenging Env. type for the experiments. The results are illustrated in Figure~\ref{fig:rt_envhardness}.a, showing that success-rate reaches 100\% as the runtime grows.

\subsubsection{Results with Different Goal Configurations}
\label{sec:dist_gconf}

\begin{figure}[b]
  \centering
  \includegraphics[scale=0.4]{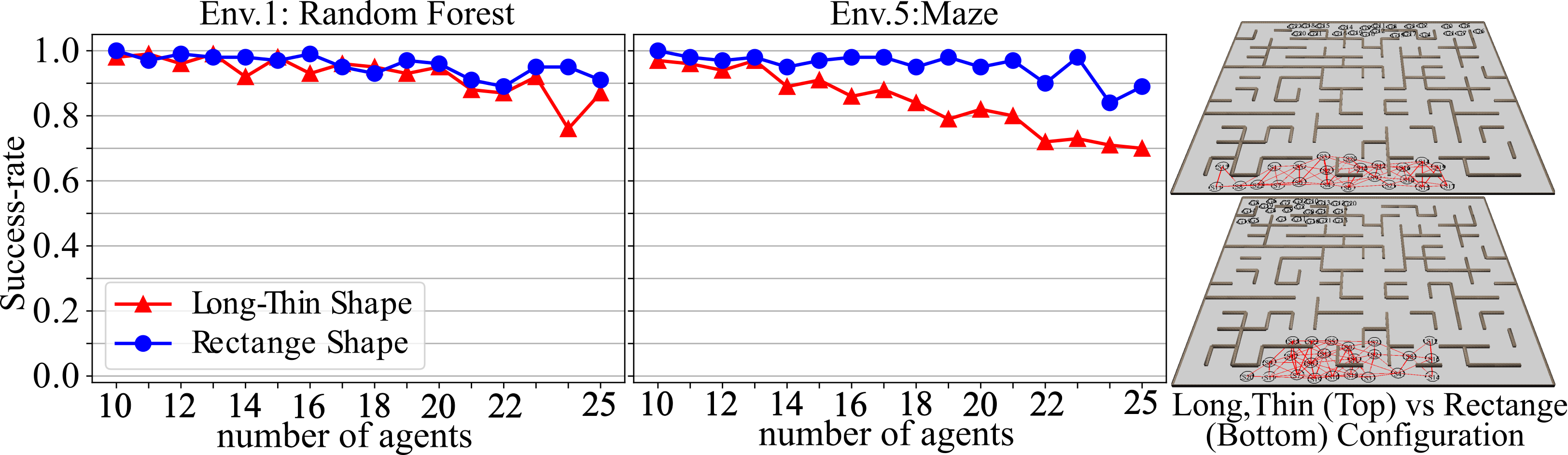}
  \caption{\textbf{Long-Thin Shape (Top) vs Rectangle Shape (Bottom) Goal Configuration.}}
  \Description{We evaluate the planner \textsc{apedl} on two goal configurations: Long-Thin and Rectangle Shapes. The Long-Thin Shape of goal configuration makes the planning challenging. }
  \label{fig:env-goaldist}
\end{figure}

The goal configuration distribution impacts the planner's performance, especially when managing diverse agent movement. A thin, long goal distribution increases runtime as the planner must search longer for a suitable planning order. We run \planner{} with an alternate goal configuration distributions: long and thin versus rectangular (Figure~\ref{fig:env-goaldist}) to test its robustness.
The \gls{DL} technique makes the \planner{} robust in handling diverse movement directions across most environments (similar to Random Forest Env.-Figure~\ref{fig:env-goaldist}.left). However, in Maze environments with narrow passages and only one way to go, frequent collisions and path modifications significantly increase difficulty of the planning problem, reducing planning performance somewhat.

\begin{figure}[bt]
  \centering
  \includegraphics[width=1.0\textwidth]{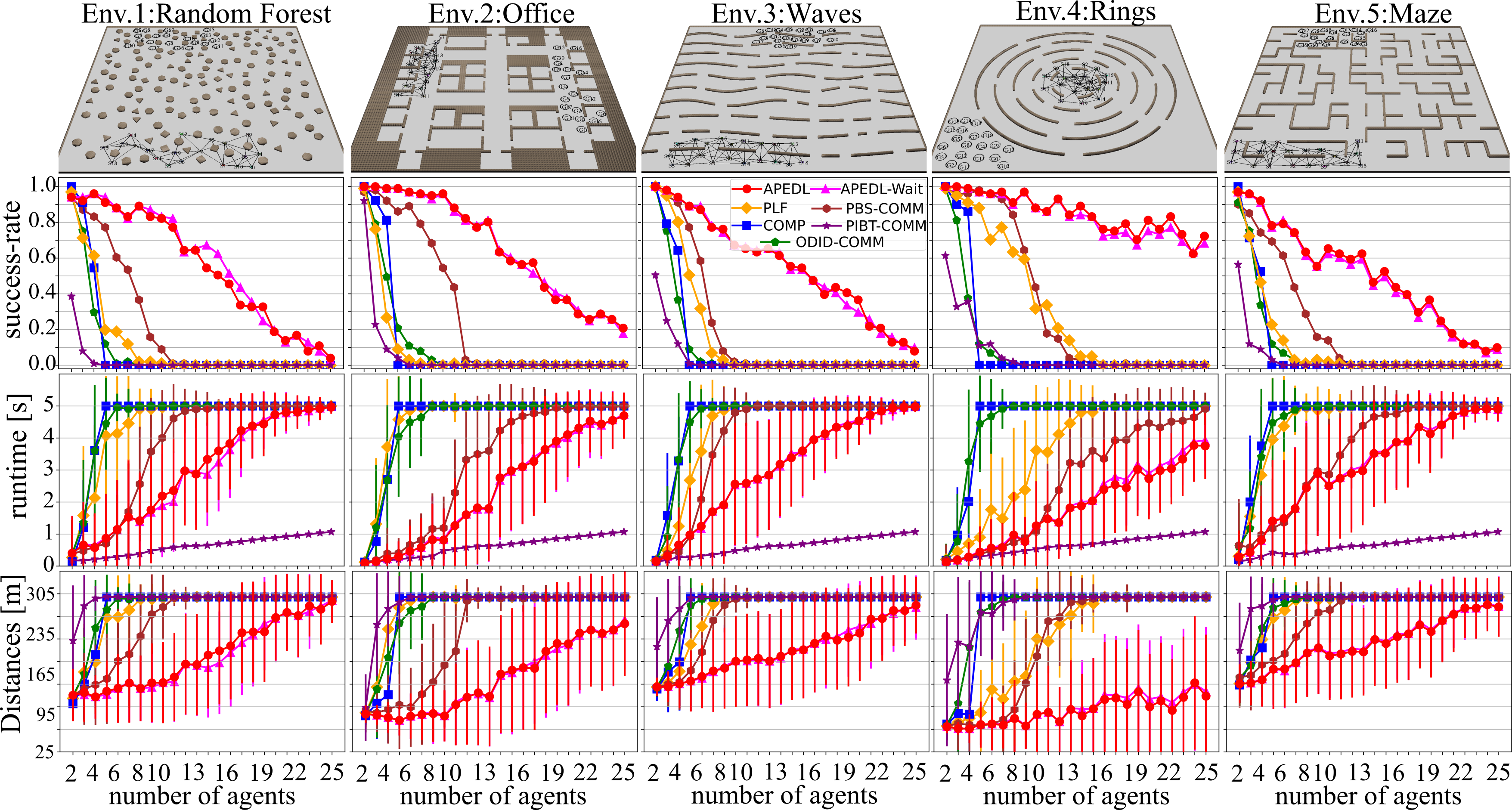}
  \caption{\textbf{Success-rate, Runtime, and Travel Distances of our proposed planner: \planner{} with \gls{PLF}, \gls{COMP}, \gls{ODIDCOMM}, \gls{PIBTCOMM}, and \gls{PBSCOMM} under \gls{LOS} constraint}. 
  The \gls{LOS} constraint poses significant challenges for the planners to compute paths for agents. The results highlight the relative effectiveness of our proposed techniques: \textit{adaptive path expansion} and \textit{dynamic leading}.}
  \Description{The planner \textsc{apedl} is evaluated under Line-of-Sight constraint.}
  \label{fig:result-LOS}
\end{figure}

\subsection{Results under Line-of-Sight Communication Constraint}
\label{sec:los_results}
We test the performance of our \planner{} planner under \gls{LOS} constraint, which poses significant challenges due to the reason: maintaining the \gls{LOS} constraint is difficult in obstacles-rich environments.
In experiments using the same environments and instances as those with \gls{LCR} constraint, we replace \gls{LCR} with \gls{LOS} constraint, with results presented in Figure~\ref{fig:result-LOS}.

Under the \gls{LOS} constraint, the \planner{} planner still outperform baseline methods, demonstrating the robustness of our proposed techniques: \textit{adaptive path expansion} and \textit{dynamic leading}. The \planner{} planner achieves superior performance, successfully computing paths for 11--12 agents in Office and Rings environments and 4--6 agents in Random Forest, Waves, and Maze environments. 

Maintaining the \gls{LOS} communication in long, narrow passages or obstacle-rich environments is challenging, as a leader pursuing its goal (e.g., entering a room or navigating around obstacles) may inadvertently break communication with followers.


\section{Completeness Analysis}
\label{sec:completeAna}
\begin{figure}[t]
  \centering
  \includegraphics[width=0.44\linewidth]{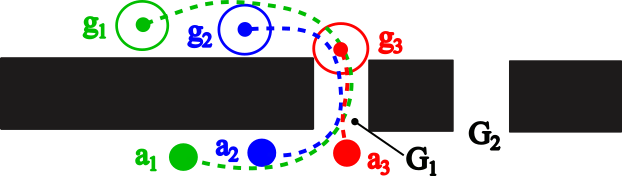}
  \caption{\textbf{A scenario demonstrating the incompleteness of planning framework.} Due to the greedy nature of \saplannerdl{} (find the shortest path), all agents attempt to go through $G_1$ without trying $G_2$ and cannot all reach their respective goals.}
  \Description{The \textsc{apedl} is still incomplete in this domain.}
  \label{fig:complete_break}
\end{figure}

Though our planner represents an advance over leader-follower and \gls{PBS} planners by overcoming their reliance on planning order that results in their incompleteness in solving the \gls{MAT3C} problem, the \planner{} planners is also \textit{incomplete} in this domain. 
This is a consequence of the greedy nature of the single agent planner (\saplannerdl), which always immediately returns upon finding a valid single agent path. We illustrate a scenario in Figure~\ref{fig:complete_break} 
that \planner{} planner fails to solve: though the solution requires that agent $a_3$ selects a longer path to allow other agents to pass by, the \saplannerdl{} planner returns only the shortest path and so the team becomes stuck.

Future work could explore extending our planning framework to achieve completeness. For this to be possible, the \saplannerdl{} single-agent planner would need to be extended to eventually generate all possible paths when expanding each agent, a modification that would require careful consideration and detailed study so that it would not dramatically slow planning performance.
In this research, our focus is on developing a fast and practical planner that can efficiently handle the \malcr{} problem, outperforming existing approaches. We plan to develop a complete version of the \planner{} planner in future research.

\section{Conclusion} \label{sec:conclusion}

This paper presents a novel two-level planning framework, \planner{}, to address the \malcr{} problem. 
The core advancements of our approach are twofold: (1) at the high level, we deploy an \textit{adaptive path expansion} technique to expand and refine agent paths multiple times to enable the planner to compute paths with arbitrary start and goal configurations, where state-of-the-art \gls{MAPF} planners such as \gls{PBS}, \gls{PIBT}, \textsc{od-id} fail; 
and (2) at the low level, single-agent pathfinding with \textit{dynamic leading} enables the dynamic reselection of the leading agent during single-agent path expansions whenever progress stalls, helping our planner to outperform state-of-the-art priority-based search approaches.
We have tested our planner in multiple environments with features such as 
obstacles-richness, narrow passages, non-convex spaces, and long hallways. The results demonstrate the robustness of our planning framework \planner{} capable of planning up to 25 agents under the \gls{LCR} constraint, and up to 11--12 agents in Rings and Office environments and up to 3--10 agents in other environments under the \gls{LOS} constraint within 5 seconds of runtime.
Using the \planner{} as a step-stone, in future work, we hope to extend these contributions to enable a version of the planner that has the completeness property.
There are also potential performance improvements to be gained through the design of a new heuristic that improves the expansion of the team communication tree.
Furthermore, we plan to extend the framework to \added{a continuous action space} to better support robot teams with rich kinodynamic constraints\added{, our work a step towards communication-constrained planning more broadly}. \added{In this setting, \textsc{sipp} is a promising choice for the low-level planner because it maintains a smaller state space by reasoning over time interval rather than time step, as in A*.
Incorporating continuous actions and kinodyamic constraints would bring \planner{} a step closer to real-world applications.}

\begin{acks}
The work by E. Plaku is supported by (while serving at) the U.S. National Science Foundation. Any opinion, findings, and conclusions or recommendations expressed in this material are those of the authors, and do not necessarily reflect the views of the National Science Foundation.

This material is based upon work supported by the National Science Foundation under Grant No. 2232733. G. J. Stein acknowledges support from this grant. G. J. Stein is also supported by Army Research Lab under award W911NF2520011.
\end{acks}
\printbibliography
\end{document}